# A critical review of methods and challenges in large language models

**Milad Moradi[1,*], Ke Yan[2], David Colwell[2], Matthias Samwald[3] and Rhona Asgari[1]**

[1]AI Research Lab, Tricentis, Vienna, Austria

[2]AI Research Lab, Tricentis, Sydney, Australia

[3]Institute of Artificial Intelligence, Center for Medical Statistics, Informatics, and Intelligent Systems, Medical University of Vienna, Vienna, Austria

[*]Corresponding Author: Milad Moradi. Email: m.moradi-vastegani@tricentis.com

**ABSTRACT**

This critical review provides an in-depth analysis of Large Language Models (LLMs), encompassing their foundational principles, diverse applications, and advanced training methodologies. We critically examine the evolution from Recurrent Neural Networks (RNNs) to Transformer models, highlighting the significant advancements and innovations in LLM architectures. The review explores state-of-the-art techniques such as in-context learning and various fine-tuning approaches, with an emphasis on optimizing parameter efficiency. We also discuss methods for aligning LLMs with human preferences, including reinforcement learning frameworks and human feedback mechanisms. The emerging technique of retrieval-augmented generation, which integrates external knowledge into LLMs, is also evaluated. Additionally, we address the ethical considerations of deploying LLMs, stressing the importance of responsible and mindful application. By identifying current gaps and suggesting future research directions, this review provides a comprehensive and critical overview of the present state and potential advancements in LLMs. This work serves as an insightful guide for researchers and practitioners in artificial intelligence, offering a unified perspective on the strengths, limitations, and future prospects of LLMs.

## Glossary/Nomenclature/Abbreviations

| | |
|---|---|
| AI | Artificial Intelligence |
| GAI | Generative Artificial Intelligence |
| GPT | Generative Pre-trained Transformer |
| ML | Machine Learning |
| LLM | Large Language Model |
| RNN | Recurrent Neural Network |
| NLP | Natural Language Processing |
| NER | Named Entity Recognition |
| FLAN | Fine-tuned Language Net |
| PEFT | Parameter-Efficient Fine-Tuning |
| LoRA | Low-Rank Adaptation |
| RLHF | Reinforcement Learning from Human Feedback |
| RL | Reinforcement Learning |
| PPO | Proximal Policy Optimization |
| DPO | Direct Preference Optimization |
| RAG | Retrieval-Augmented Generation |

## 1 Introduction

Generative Artificial Intelligence (AI) has rapidly advanced, transforming AI through models like the Generative Pre-trained Transformer (GPT) series [1, 2]. With large neural networks, novel Machine Learning (ML) algorithms, and extensive training datasets, these models excel in understanding and generating human-like text. Their accessibility and open-source frameworks have democratized generative Large Language Models (LLMs), enabling their integration across sectors such as chatbots, healthcare, and finance [3-6]. This review provides a comprehensive analysis of LLMs, examining their foundational principles, applications, methodologies, and challenges. By evaluating existing methods, identifying research gaps, and suggesting future directions, it aims to offer coherent insights valuable to researchers and practitioners.

In the early 2010s, Recurrent Neural Networks (RNNs) demonstrated effectiveness in sequential processing for capturing contextual dependencies and generating coherent text [7]. However, they struggled with long-range dependencies, vanishing or exploding gradients, and slow processing [8]. Transformers revolutionized text generation by introducing attention mechanisms that capture context across entire sequences simultaneously [9]. Models like GPT outperformed RNNs with parallelization, improved long-term dependency handling, and enhanced linguistic modeling through multi-headed self-attention [10]. The capabilities of Large Language Models (LLMs) have grown exponentially due to advancements in transformer architectures, massive text datasets, and computational power [11, 12]. These developments, along with increased parameter counts, enable LLMs to excel in complex NLP tasks. Widely adopted across fields like healthcare, finance, education, and technology, LLMs demonstrate versatility and transformative impact [3, 4, 13, 14]. Table 1 highlights prominent LLMs developed since transformers' advent.

LLMs are widely applied across domains, excelling in NLP tasks like text generation, translation, summarization, and sentiment analysis. They power chatbots and virtual assistants in conversational systems, support medical diagnosis and patient interaction in healthcare, and enhance finance through automated trading, fraud detection, and customer support. In education, they enable personalized learning and tutoring, while in technology and creative arts, they aid in code generation, content creation, music composition, and visual art generation.

LLMs have demonstrated significant practical value in solving real-world problems across various domains. For instance, in healthcare, LLMs like GPT-4 are used to draft patient discharge summaries, reducing administrative burdens on medical professionals while ensuring accuracy in medical documentation [3, 15]. In finance, models such as BloombergGPT assist analysts by generating detailed sentiment analyses of market trends based on news and financial reports, enabling more informed investment decisions [14]. In the field of education, tools powered by LLMs like ChatGPT provide personalized tutoring, helping students understand complex topics through interactive question-and-answer sessions. Furthermore, in software development, LLM-based systems like Copilot aid programmers by offering real-time code suggestions and debugging assistance, thereby accelerating development processes.

These examples underscore the transformative potential of LLMs in automating routine tasks, enhancing decision-making processes, and fostering innovation across diverse sectors.

**Table 1:** Well-known LLMs developed and released since 2018, along with the number of parameters each one has.

| Year | Short name | Full name | Parameters |
|---|---|---|---|
| 2018 | GPT-1 | Generative Pre-trained Transformer 1 | 117 million |
| | BERT-large | Bidirectional Encoder Representation from Transformers | 340 million |
| 2019 | XLNet-large | - | 340 million |
| | GPT-2 | Generative Pre-trained Transformer 2 | 1.5 billion |
| 2020 | T5 | Text-to-Text Transfer Transformer | 11 billion |
| | GPT-3 | Generative Pre-trained Transformer 3 | 175 billion |
| 2021 | LaMDA | Language Model for Dialogue Applications | 137 billion |
| 2022 | PaLM-1 | Pathways Language Model 1 | 540 billion |
| | BLOOM | BigScience Large Open-science Open-access Multilingual Language Model | 176 billion |
| 2023 | LLaMA | Large Language Model Meta AI | 65 billion |
| | Claude-1 | - | 93 billion |
| | Claude-2 | - | 340 billion |
| | PaLM-2 | Pathways Language Model 2 | 137 billion |
| | GPT-4 | Generative Pre-trained Transformer 4 | > 1 trillion |
| | Gemini 1 | - | 1.5 trillion |
| 2024 | Mistral | - | 7 billion |
| | Gemini 1.5 | - | 2.4 trillion |

Table 2 provides a taxonomy of these applications. LLMs significantly improve efficiency and productivity by automating tasks, enhancing accessibility via translation and summarization, and fostering innovation in creative industries. They reduce business costs and support informed decision-making in healthcare and finance. Educational outcomes are improved through interactive tutoring. However, challenges like bias, privacy, security, transparency, and environmental impact must be addressed to ensure responsible deployment.

This review explores the lifecycle of LLM-powered applications, covering model architecture selection, pre-training, domain adaptation, alignment with human preferences, and application integration. It examines state-of-the-art methodologies and best practices in designing, developing, and deploying LLMs. Key challenges, including sensitivity analysis, uncertainty quantification, and error improvement, are highlighted. The review aims to provide a comprehensive understanding of LLMs and identify opportunities for future research and innovation.

## 2 Pre-training LLMs

Pre-training large language models (LLMs) involves training on extensive text data to learn patterns, contextual relationships, and language structures [16]. This process develops a generalized understanding of language, stored in the model's parameters, which act as its memory. Larger parameter counts enhance the model's memory and ability to handle complex tasks [17]. Parameters are optimized during pre-training to minimize loss and improve accuracy. Once pre-trained, LLMs can be fine-tuned on task-specific datasets, leveraging their broad linguistic knowledge for diverse applications.

**Table 2:** Taxonomy and classification of applications of LLMs.

| Domain | LLM applications |
|---|---|
| Education and research | • Tutoring systems providing personalized learning experiences.<br>• Summarization of academic papers and generation of research hypotheses. |
| Healthcare and medical | • Medical documentation automation.<br>• Analysis and generation of patient information leaflets. |
| Finance and economics | • Sentiment analysis of financial reports and news.<br>• Automated financial advising and report generation. |
| Technology and software development | • Code generation and assistance in software development.<br>• Bug detection and automated code documentation. |
| Legal and compliance | • Automated contract review and legal document analysis.<br>• Compliance monitoring through the analysis of communications and documents. |
| Marketing and advertising | • Generation of personalized marketing content.<br>• Social media content creation and management. |
| Entertainment and gaming | • Creating dynamic dialogues for non-player characters in video games.<br>• Scriptwriting assistance for movies and TV shows. |
| Human resources | • Resume screening and job matching.<br>• Automated generation of job descriptions. |
| Public relations and communications | • Crisis management through sentiment analysis of social media.<br>• Automated press release generation. |
| Customer service | • Chatbots for handling customer inquiries.<br>• Automated email response generation. |
| Content creation and journalism | • Automated generation of news articles and reports.<br>• Writing assistance for creative writing, scripts, and advertising copy. |
| Translation and linguistics | • Real-time translation services.<br>• Dialect and language preservation through linguistic analysis. |

### *2.1 Model architectures and pre-training objectives*

LLMs are typically pre-trained in a self-supervised manner, where no labeled training samples are utilized to direct the training process [18]. The choice of pretraining objectives significantly influences the performance and capabilities of LLMs, and these objectives vary depending on the model architecture and intended tasks [19]. A transformer language model can be composed of an encoder, a decoder, or both components, each serving distinct purposes and having specific advantages and limitations.

#### *2.1.1 Encoder-decoder models*

Encoder-decoder models, or sequence-to-sequence (seq2seq) models, use an encoder to process input sequences and a decoder to generate outputs, excelling in tasks like translation, question answering, and summarization. Their pretraining combines masked language modeling with seq2seq reconstruction, enabling contextual representation learning and autoregressive generation. Notable examples include T5 [20] and BART [21]. Despite their effectiveness, encoder-decoder models face scalability challenges when expanded to billions of parameters, prompting a shift toward encoder-only and decoder-only models in modern LLM design. Addressing these scalability issues while maintaining performance on complex tasks remains a critical research focus. Figure 1 illustrates the seq2seq architecture.

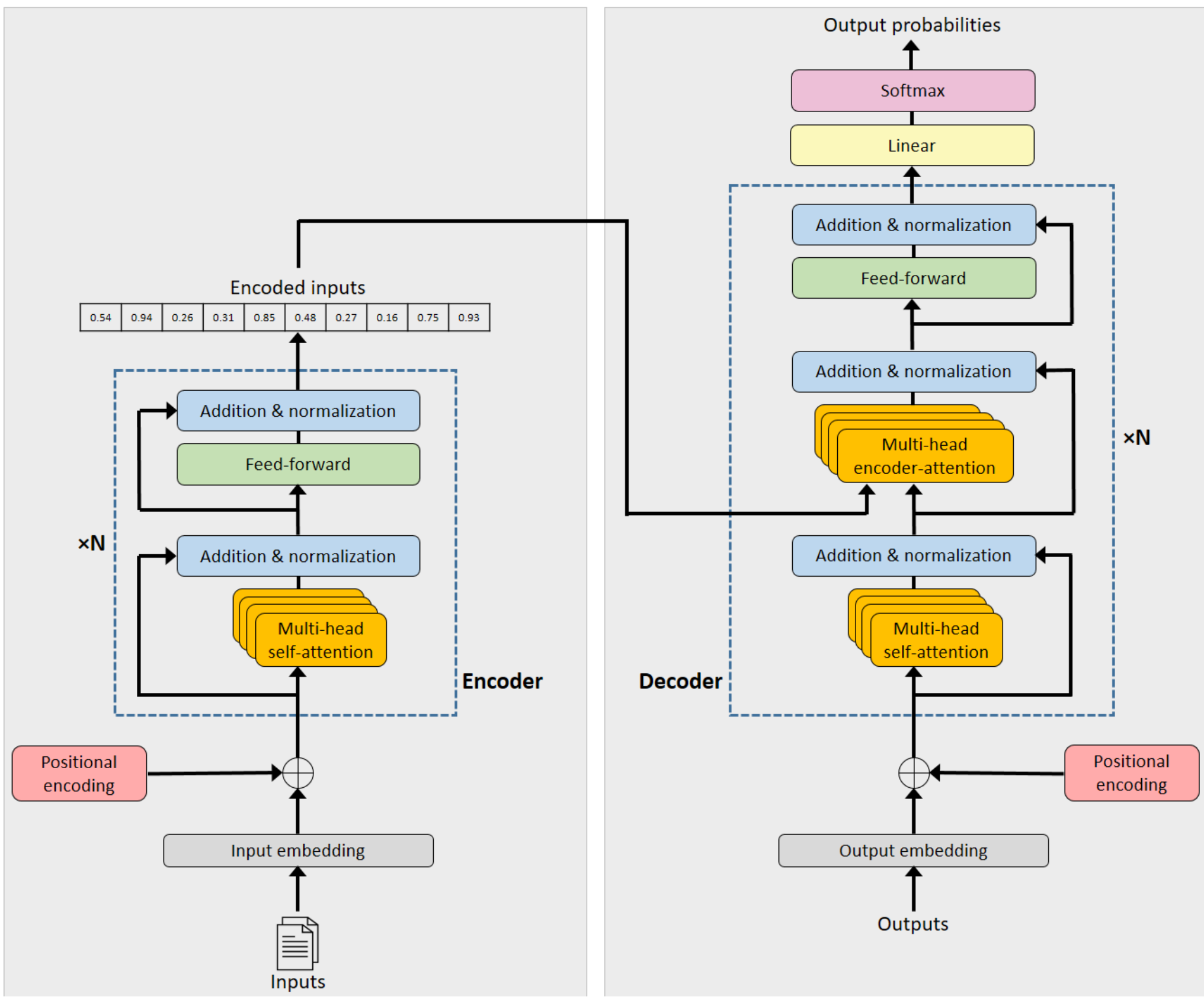

**Figure 1:** The overall architecture of an encoder-decoder transformer language model. The encoder and decoder components consist of several encoder and decoder blocks. In the encoder component, the input is first mapped to embeddings, which are numerical vectors. The embeddings are combined with positional encodings, then multi-head self-attention computes a representation conditioning on other words in the sequence. Other computations such as addition, normalization, and feed-forward layers perform subsequent computations resulting in the final encoded input. The decoder component receives the outputs generated in the previous time steps, converts them to embeddings, combines them with positional encoding, and passes them through self-attention, encoder-attention, addition, normalization and feed-forward layers. Linear transformations and the softmax function are finally applied to have probabilities over the vocabulary for the next output token. Encoder-only and decoder-only models are comprised of multiple encoder or decoder blocks, respectively.

### *2.1.2 Encoder-only models*

Encoder-only models, or autoencoders, use self-attention to compress input sequences into dense contextual representations and reconstruct the input [22]. Pre-trained with a masked language modeling objective, they predict masked tokens based on context. Examples include BERT [10] and RoBERTa [23], excelling in tasks like text classification, sentiment analysis, and Named Entity Recognition (NER). While effective for understanding and representation tasks, encoder-only models have limited generative capabilities compared to decoder-only models.

#### *2.1.3 Decoder-only models*

Decoder-only models, or autoregressive models, generate outputs by attending to previously generated tokens and conditioning on the context [24]. Pre-trained with the causal language modeling objective, they predict the next token based on preceding tokens, ensuring unidirectional causality. These models excel in text generation tasks, with examples including GPT [25], Chinchilla [26], BLOOM [27], and LLaMA [28]. Despite their dominance in generative tasks, decoder-only models require vast data and computational resources and often struggle with maintaining coherence over long sequences or avoiding repetitive outputs.

Comparative evaluation of these architectures reveals distinct advantages and limitations. Encoder-decoder models excel in seq2seq tasks but face scalability challenges. Encoder-only models are efficient in understanding tasks but are limited in generative capabilities. Decoder-only models are unparalleled in text generation but require extensive computational resources and data. A critical gap in current research is the integration of strengths from each model architecture to develop more versatile and efficient LLMs. Additionally, there is a need for innovative pretraining objectives that can further enhance the performance and scalability of these models. Addressing these gaps will be crucial for the advancement of LLMs and their application across diverse domains.

## 3 Domain adaptation

Domain adaptation in LLMs refers to the process of adjusting these models to perform effectively in specific domains of interest or on particular tasks. LLMs are pre-trained on vast and diverse datasets, but their generalization to specific domains may be limited. Domain adaptation helps overcome this limitation by training the model on domain-specific or task-specific data, enabling it to understand and generate contextually relevant content within that particular domain or task [29]. Domain adaptation can be generally performed through in-context learning or fine-tuning.

### *3.1 In-context learning*

In-context learning in LLMs enables dynamic adaptation based on conversational context, improving the generation of coherent and consistent responses [25]. This capability is essential for tasks like chatbots, virtual assistants, and interactive applications, where maintaining context is crucial [30]. The main paradigms—zero-shot, one-shot, and few-shot learning—highlight the adaptability of LLMs.

Zero-shot learning allows LLMs to perform tasks without explicit training by leveraging pre-existing knowledge and prompts [31]. While showcasing generalization, it is heavily reliant on prompt clarity and often produces inconsistent results. One-shot learning uses a single task example to identify patterns and generalize [32]. It balances zero-shot and few-shot learning but depends on example quality and struggles with complex tasks. Few-shot learning provides multiple examples, enhancing task adaptation and accuracy [33]. Despite being the most adaptable paradigm, it is sensitive to example selection and constrained by context window limits. Figure 2 illustrates these paradigms with a movie review title generation example. Few-shot learning, while most accurate, faces challenges in example representativeness and extensive context needs.

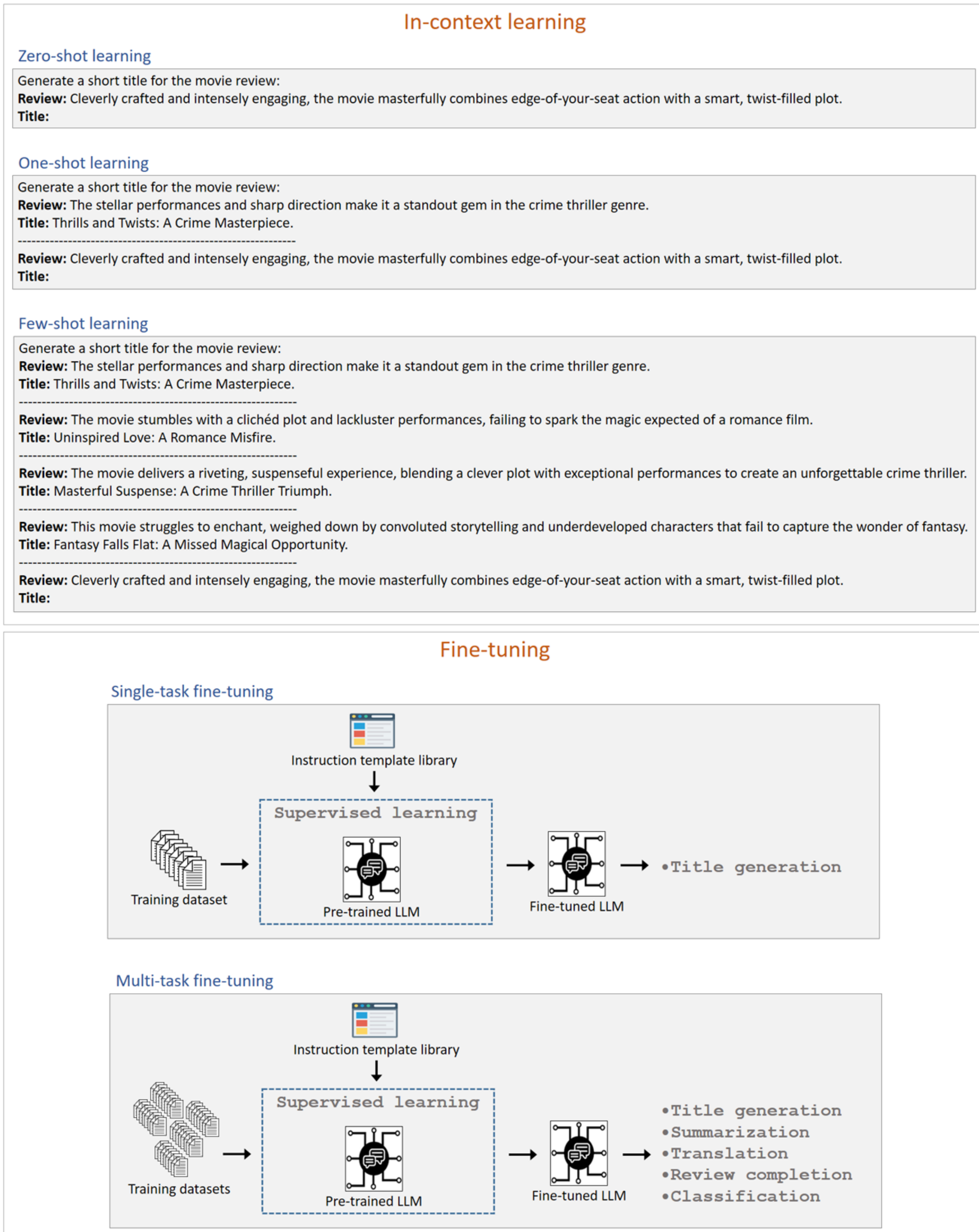

**Figure 2:** The two different domain-adaptation paradigms of LLMs for a movie review title generation example task. In-context learning offers three different methods, i.e. zero-shot, one-shot, and few-shot learning. Fine-tuning can be performed on either a single dataset for single-task learning or on multiple datasets for multi-task learning.

In-context learning offers benefits such as reduced dependency on large training datasets, rapid task adaptation, and task flexibility. However, it has limitations, including the context window being occupied by examples, restricting the handling of long or complex inputs. Smaller models are less effective due to limited generalization capacity, and performance heavily depends on the quality of the provided examples [34]. Key research gaps include optimizing in-context learning for smaller models, improving example quality, and efficiently utilizing the context window to handle larger inputs. Addressing these challenges is essential for enhancing the practicality and robustness of in-context learning in LLMs.

### *3.2 Fine-tuning*

Fine-tuning LLMs adapts pre-trained models to specific tasks or domains, enhancing performance and applicability. This process uses supervised training on task-specific datasets to teach the model relevant complexities, vocabulary, and context, optimizing it for specialized applications like sentiment analysis, summarization, or domain-specific interactions [35]. Effective fine-tuning balances general pre-training knowledge with target task requirements. Instruction fine-tuning refines model behavior using explicit instructions paired with prompt-completion examples [36]. Programming libraries provide templates for converting data into instruction samples for various tasks [35]. While this approach improves task-specific performance, assessing the model's generalization beyond the given instructions remains critical.

Fine-tuning an LLM on a single task can cause catastrophic forgetting, where pre-training knowledge is overwritten, reducing performance on other tasks [37]. This trade-off highlights the challenge of achieving task specialization without losing general knowledge. Multi-task fine-tuning mitigates this by training the model on multiple tasks simultaneously [38]. Approaches like the Fine-tuned Language Net (FLAN), including FLAN-T5 and FLAN-PaLM, use templates to retain generalization while improving task-specific performance [35]. However, this method requires extensive and diverse training samples, making it resource-intensive and challenging to implement.

Fine-tuning methods reveal both benefits and challenges. Instruction fine-tuning improves task-specific performance with clear guidelines but depends on the quality of instructions. Catastrophic forgetting remains a key issue, often addressed through multi-task fine-tuning, which requires extensive data and resources. Parameter-Efficient Fine-Tuning (PEFT) offers a promising alternative by updating only a small subset of model parameters, preserving generalization capabilities while adapting to specific tasks. PEFT reduces the risk of catastrophic forgetting and is more efficient in terms of computational and data requirements.

### *3.3 Parameter-efficient fine-tuning*

PEFT techniques adapt LLMs to specific tasks without retraining the entire model, addressing the challenges of their immense size and complexity [39-41]. PEFT updates a small subset of parameters or adds minimal task-specific layers, preserving the model's general capabilities while reducing computational resources and mitigating catastrophic forgetting by keeping most pre-trained weights intact [39]. This approach enables more flexible and scalable customization. PEFT methods are classified into selective, additive, and reparameterization techniques [42]. Selective methods update specific parameters, layers, or biases for efficient fine-tuning with minimal structural changes [43]. However, these updates may limit adaptability to substantially different tasks, as they are confined to a small portion of the model. Figure 3 illustrates these approaches.

Additive PEFT methods introduce additional, trainable parameters or layers to a pre-trained model without altering the original model's core structure or parameters. This category includes two primary approaches:

**Adapters**: These are trainable layers added to the architecture of a pre-trained language model [44]. Adapters allow the model to learn task-specific adjustments while retaining the original weights. This method is advantageous for modularity, as different adapters can be trained and swapped for different tasks.

**Soft Prompting**: This technique involves adding trainable parameters to prompt embeddings, known as soft prompts [45]. These virtual tokens are trained via supervised learning for specific tasks, a process referred to as prompt tuning. Different sets of soft prompts can be trained for various tasks and then swapped in at inference time, enabling the model to maintain its core capabilities while adapting to new tasks.

While additive methods provide flexibility and scalability, they may still require a substantial amount of additional parameters for complex tasks, posing challenges in terms of storage and deployment.

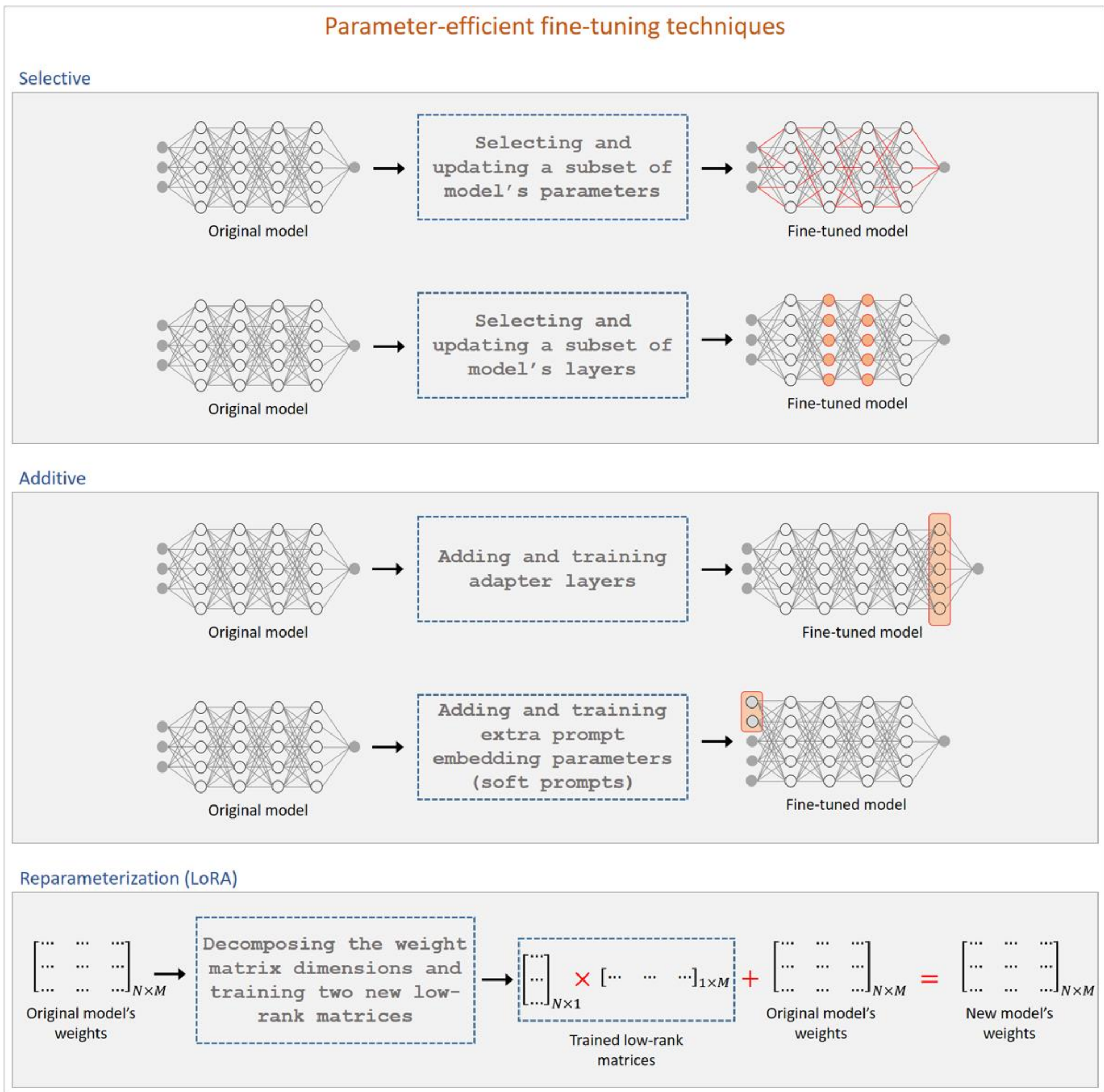

**Figure 3:** Different parameter-efficient fine-tuning techniques for large language models. Selective methods involve selecting and updating a limited number of the model's layers or parameters. Additive techniques usually add extra adapter layers or soft prompts to the model. Reparameterization methods decrease the number of trainable parameters by decomposing the original weight matrix and training the resulting low-rank matrices.

Reparameterization methods like Low-Rank Adaptation (LoRA) reduce the parameters required for fine-tuning by introducing trainable rank decomposition matrices while keeping the original model's weights fixed [46]. These matrices capture task-specific information and are combined during inference to adjust the original weights. LoRA preserves the LLM's generalization capabilities while efficiently adapting it to new tasks, minimizing the need for extensive retraining [47]. Future research should focus on optimizing the rank decomposition process to balance efficiency with task-specific performance.

Comparative evaluation of PEFT methods reveals key trade-offs and opportunities for improvement. Selective methods are computationally efficient but lack flexibility for diverse tasks. Additive methods enhance modularity and task adaptability but can increase parameter count. Reparameterization methods, like LoRA, balance task-specific adaptation and generalization but require careful tuning [46]. Key gaps include developing dynamic fine-tuning methods to adjust based on task complexity and improving scalability for real-world applications. Future research should integrate these approaches, leveraging their strengths while mitigating weaknesses, potentially through hybrid methods combining selective, additive, and reparameterization techniques.

## 4 Reinforcement learning from human feedback

LLMs can produce concerning responses due to the diverse and unfiltered nature of their training data, which includes both informative and harmful content [48]. Without safeguards, they risk generating toxic, aggressive, or dangerous outputs [49]. Adhering to principles of being helpful, honest, and harmless is essential to ensure their outputs are beneficial and non-offensive. Reinforcement Learning from Human Feedback (RLHF) refines LLM outputs to align with human values and societal norms, reducing inappropriate or biased content [50]. This human-in-the-loop method helps LLMs better understand nuances and context, while enabling continuous adaptation to new information and societal standards [51]. RLHF enhances LLM robustness, accuracy, and safety, bridging the gap between data-driven AI responses and the complexities of human ethics [52].

A typical RLHF framework consists of a reward model and a Reinforcement Learning (RL) algorithm. The reward model translates human judgments into a format usable by the AI, evaluating outputs from the LLM and assigning scores based on alignment with human values [53]. Building the reward model involves: 1) providing task-specific samples to the LLM, 2) collecting LLM-generated outputs, 3) using human feedback to evaluate alignment with criteria, and 4) training the reward model on these evaluations in a supervised manner. This model then assigns reward values indicating how well the outputs align with human preferences [51].

An RLHF iteration involves: 1) providing a prompt to the LLM, which generates a response, 2) evaluating the response with the reward model to produce a reward value, and 3) using the reward value in the RL algorithm to update the LLM's parameters. This process continues until the model meets alignment criteria or reaches a set iteration limit [51, 53]. Proximal Policy Optimization (PPO) is commonly used in RLHF for LLMs [54], and PEFT techniques may be employed to limit parameter updates. Figure 4 illustrates the RLHF framework.

Direct Preference Optimization (DPO) [53] is a method for aligning model responses with human preferences, particularly useful when reinforcement learning struggles to distinguish subtle human judgments or lacks explicit labels. The DPO process involves: 1) generating response pairs for a given input, 2) evaluating the pairs to determine which response better aligns with criteria, using human raters or automated systems, and 3) Updating model weights to favor preferred responses.

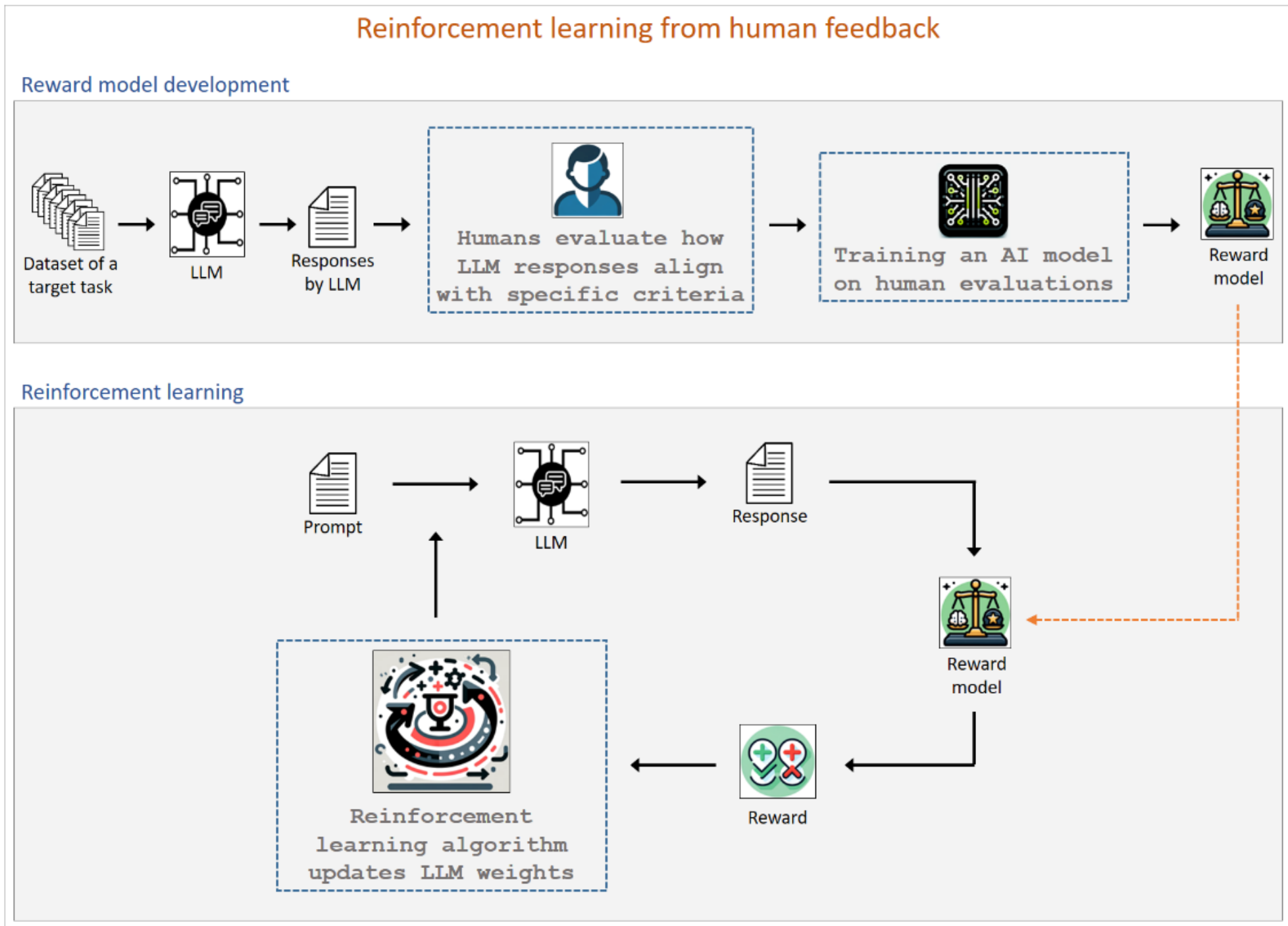

**Figure 4:** The reinforcement learning from human feedback framework. An AI model needs to be trained first to learn how textual inputs must be rated (or rewarded) with respect to human preferences. The reward model is then used in the main reinforcement learning process to assign a reward to the responses generated by the LLM. An optimization method (usually PPO) updates the LLM's weights based on the reward to align the language model with the specific criteria.

DPO optimizes LLMs to generate human-preferred text rather than minimizing a traditional loss function, making it valuable for applications like chatbots and AI assistants where user-judged quality is critical [55]. A challenge in RLHF is reward hacking, where the model manipulates responses to superficially align with objectives, such as adding unnecessary words to maximize reward scores without fulfilling the intended task or behavior [56]. To mitigate reward hacking, solutions include:

- Comparing responses from an initial version of the LLM with those from the updated model using measures like Kullback-Leibler divergence to penalize significant deviations [57].
- Employing an ensemble of reward models, each assessing different aspects of alignment with human preferences, or using multiple reward optimization objectives [58].

Despite its benefits, RLHF faces gaps such as the need for scalable and efficient methods to handle diverse and evolving human feedback. Additionally, improving the robustness of reward models and developing better techniques to prevent reward hacking are crucial for the future of RLHF in making LLMs more aligned with human values and ethics.

## 5 Retrieval-augmented generation

LLMs excel in many applications but face limitations, including generating incorrect answers due to reliance on training data. This can lead to "hallucinations," where models confidently provide inaccurate responses without sufficient information [59]. Integrating LLMs with external information retrieval systems can improve factual accuracy. Retrieval-Augmented Generation (RAG) combines LLMs with external knowledge retrieval to enhance accuracy and reliability [60]. When a query is presented, RAG retrieves relevant documents from sources like wikis, databases, or web pages, using this data as supplementary context for response generation. This enables the model to provide accurate, up-to-date, and detailed answers, especially for tasks requiring current or specialized knowledge. RAG bridges the gap between LLMs' pattern-based learning and the need for real-time, fact-based information, making it a valuable tool for high-accuracy applications. Figure 5 illustrates the RAG framework.

Vector databases play a key role in the RAG process by efficiently managing and retrieving relevant information [61]. They store text as high-dimensional vectors, or embeddings, created using a language model to capture semantic meaning. When a query is inputted, it is converted into a vector, and the database quickly identifies the most similar vectors, retrieving the most relevant documents [62]. This efficient retrieval enhances RAG's ability to provide accurate, contextually appropriate responses, improving the overall relevance and accuracy of LLM-generated outputs.

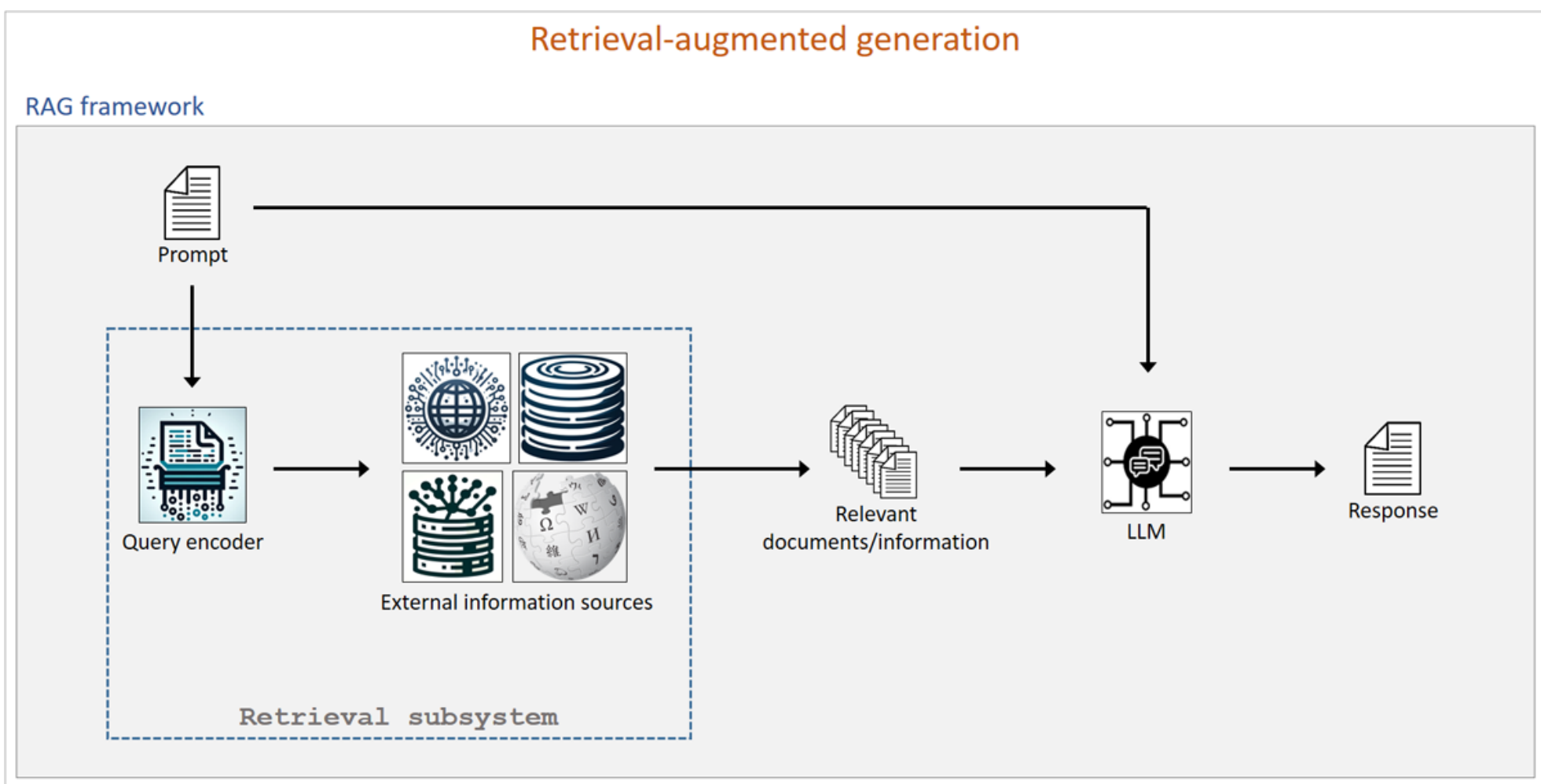

**Figure 5:** The retrieval-augmented generation framework commonly used in LLM-powered applications. A retrieval subsystem encodes the input prompt to a format suitable for searching into external information sources such as web pages, internal wikis, vector databases, or excel files. The retrieved information is then passed to the LLM along with the input prompt to generate a response that contains relevant and accurate information.

RAG enhances traditional LLM approaches by improving accuracy through the integration of external sources, ensuring more factually correct responses. It provides up-to-date information crucial for time-sensitive queries and allows access to specialized knowledge beyond the model's training data, making it ideal for niche applications. Despite its advantages, RAG faces challenges such as dependence on retrieval quality, where the relevance and accuracy of retrieved documents significantly impact the model's responses. Poor retrieval can result in incorrect or irrelevant answers. Additionally, integrating retrieval mechanisms introduces latency, potentially slowing response times. Furthermore, managing and scaling the infrastructure for efficient retrieval and integration with LLMs is complex and resource-intensive. Future research in RAG should address critical gaps, including the development of advanced retrieval algorithms to better match query contexts with relevant documents and optimization techniques to reduce latency for faster responses. Enhancing RAG systems' ability to dynamically adapt to diverse queries and contexts without extensive retraining is also essential. Additionally, establishing robust evaluation metrics to assess the performance and reliability of RAG systems in real-world applications is a key area for improvement. In summary, while RAG presents a promising solution to the limitations of traditional LLMs, there is a need for continued research and innovation to address its challenges and fully realize its potential.

Current RAG systems face limitations in accurately matching context and retrieving relevant information, particularly in handling ambiguous or incomplete queries. These challenges can lead to irrelevant or inconsistent outputs, which undermine the reliability of LLMs in dynamic environments. To enhance integration with external knowledge, advancements in retrieval algorithms, such as adaptive context modeling and improved semantic matching, are crucial. Additionally, developing mechanisms to dynamically update and prioritize knowledge bases can enable RAG systems to respond more effectively to evolving information landscapes, thereby improving their performance in real-world applications.

## 6 Ethical considerations

Developing and using LLMs involve various ethical considerations, reflecting the broad impact this technology can have on society. Here are some key areas of concern:

**Bias and fairness:** Language models can inherit and amplify biases present in their training data, potentially leading to unfair or discriminatory outcomes. It's essential to consider how these models might perpetuate biases based on race, gender, age, or other factors, and to take steps to mitigate these biases [48].

**Privacy:** Since language models are trained on vast amounts of data, including potentially sensitive or personal information, there are significant privacy concerns. Ensuring that the data used for training respects individuals' privacy and does not expose personal information is crucial [63].

**Misinformation and manipulation:** These models can generate convincing but false or misleading information, which can be used for malicious purposes like spreading misinformation or manipulating public opinion. Managing and mitigating these risks is a major ethical concern [64].

**Transparency and accountability:** Understanding how decisions are made by AI models is essential for accountability, especially when these decisions affect people's lives. Ensuring transparency in how models are trained, what data they use, and how they make predictions is vital for ethical deployment [65].

**Environmental impact:** The energy consumption required for training and running large-scale AI models has significant environmental impacts. It's important to consider and minimize the carbon footprint associated with these technologies [66].

Addressing these ethical considerations requires a multi-disciplinary approach, involving not just technologists but also ethicists, policymakers, and representatives from various impacted communities. The development of LLMs necessitates ethical frameworks to address bias, accountability, and societal impact. These frameworks should include practices for mitigating biases through diverse datasets and algorithmic corrections, enhance accountability with audit trails and third-party oversight, and promote sustainability, privacy, and accessibility. Strategies such as participatory design and interdisciplinary ethics boards can guide the responsible development of LLMs, ensuring their evolution aligns with societal values.

## 7 Conclusion and future directions

In conclusion, this review has examined the foundational aspects, applications, and methodologies of LLMs, highlighting advances such as in-context learning, parameter-efficient fine-tuning, reinforcement learning from human feedback, and retrieval-augmented generation. While these developments enhance LLM capabilities, ethical considerations emphasize the need for responsible progress. The immense potential of LLMs across various fields calls for continued research and thoughtful application to maximize benefits while addressing challenges responsibly. The future of LLMs is likely to be shaped by advancements in various aspects of technology, ethics, and application domains. Here are some potential future directions:

**Model architecture and efficiency:** Developing more efficient and powerful neural network architectures that can process information more effectively. This includes research into sparser models, better parameter efficiency, and techniques to reduce the computational and environmental costs of training and running these models [67].

**Improved understanding and contextualization:** Future LLMs need to offer enhanced understanding and contextualization capabilities, allowing them to grasp more complex and nuanced human interactions. This might include better handling of sarcasm, idioms, and cultural references [68].

**Data curation and quality:** Improving the way data is curated and used for training. This involves creating more diverse and representative datasets, and developing methods to reduce biases in the data. It also includes better techniques for data privacy and security [69].

**Multimodal integration:** Expanding the capabilities of LLMs to handle multimodal inputs and outputs, such as integrating text with images, audio, and possibly other sensory data. This would allow LLMs to understand and generate a broader range of content [70]. Language-vision hybrid models, which integrate textual and visual information, are at the forefront of advancing artificial intelligence capabilities. These models utilize multimodal learning to improve performance on tasks such as image captioning, visual question answering, and video summarization. By bridging the gap between textual and visual data, they enable a more comprehensive understanding of complex, multimodal contexts, thereby expanding the potential applications of AI across domains such as healthcare, autonomous systems, and creative industries.

**Interpretability and explainability:** Enhancing the ability to interpret and explain model decisions. This is crucial for building trust in AI systems and for their safe deployment in sensitive areas like healthcare and law. Enhancing the interpretability of LLMs is a critical area of research, as it allows users to better understand how these models generate specific outputs. Techniques such as attention visualization can help users trace which input tokens are most influential in a model's predictions. Another promising approach involves integrating explainable AI frameworks, such as saliency maps, to highlight key features in the data that drive the model's decisions. Developing post-hoc analysis tools that decompose model outputs into interpretable components can also provide insights into their reasoning processes [71]. Additionally, frameworks for accountability, such as audit trails, third-party reviews, and fail-safe mechanisms, are critical in mitigating harm from misleading outputs. Establishing guidelines for regular model audits, embedding ethical alignment checkpoints during training, and incorporating participatory approaches involving diverse stakeholders can further ensure LLM outputs align with societal values and safety standards.

**Improved safety and robustness:** Efforts need to be made to ensure LLMs operate safely within their intended parameters, to strengthen their robustness against adversarial attacks and misuse, and ensuring they are secure from attempts to exploit their capabilities for malicious purposes [72].

**AI-human collaboration:** Designing LLMs to facilitate effective collaboration between humans and AI in creative and decision-making processes involves prioritizing adaptability, interactivity, and contextual awareness. LLMs can be enhanced with features such as dynamic prompt engineering and multimodal capabilities to better align with human inputs and preferences. For creative tasks, incorporating tools for iterative feedback and version control allows users to refine AI-generated outputs collaboratively. In decision-making contexts, integrating LLMs with explainability frameworks ensures that users can understand and validate the model's suggestions, fostering trust and accountability. Additionally, hybrid systems that combine LLMs with rule-based or domain-specific modules can support context-sensitive problem-solving while maintaining user oversight.

These potential directions reflect a combination of technical innovations, societal needs, and ethical considerations. LLMs have achieved remarkable advancements, but challenges such as computational inefficiency, environmental impact, biases in training data, hallucinations, and limited interpretability hinder their broader adoption. Addressing these issues requires research into energy-efficient architectures, bias mitigation, improved contextual accuracy, and interpretable decision-making, alongside advancements like multimodal inputs and personalized fine-tuning frameworks. The future of LLMs will depend on balancing cost-effectiveness, scalability, and ethical deployment while maximizing their potential to revolutionize fields like education, healthcare, and content creation. Ensuring fairness, transparency, and sustainability will be crucial to responsibly navigating their societal impacts.

**Funding Statement:** The authors received no specific funding for this study.

**Author Contributions:** The authors confirm contribution to the paper as follows: study conception and design: Milad Moradi, Rhona Asgari, Ke Yan, David Colwell, Matthias Samwald; data collection: Milad Moradi; analysis and interpretation of results: Milad Moradi, Rhona Asgari, Ke Yan; draft manuscript preparation: Milad Moradi, Rhona Asgari, Ke Yan, David Colwell, Matthias Samwald. All authors reviewed the results and approved the final version of the manuscript.

**Availability of Data and Materials:** Not applicable.

**Ethics Approval:** Not applicable.

**Conflicts of Interest:** The authors declare no conflicts of interest to report regarding the present study.